\documentstyle[conference,named,lingmacros]{article}  

\begin{document}

\input{psfig}
\bibliographystyle{named}

\title{Experimentally Evaluating Communicative Strategies: \\
The Effect of the Task \\
\begin{small} AAAI 94, Seattle \end{small}}
\author{Marilyn A. Walker \\
Mitsubishi Electric Research Laboratories\thanks{This research
was partially funded by ARO grant DAAL03-89-C0031PRI and DARPA grant
N00014-90-J-1863 at the University of Pennsylvania and by Hewlett
Packard, U.K.}
 \\ 201 Broadway \\
Cambridge, Ma. 02139, USA \\{\tt walker@merl.com} }

\maketitle

\begin{abstract}
\begin{quote}
Effective problem solving among multiple agents requires a better
understanding of the role of communication in collaboration.  In this
paper we show that there are communicative strategies that greatly
improve the performance of resource-bounded agents, but that these
strategies are highly
sensitive to the task requirements, situation parameters and agents'
resource limitations.  We base our argument on two sources of
evidence: (1) an analysis of a corpus of 55 problem solving dialogues,
and (2) experimental simulations of collaborative problem solving
dialogues in an experimental world, Design-World, where we
parameterize task requirements, agents' resources and communicative
strategies.
\end{quote}
\end{abstract}


\section{Introduction}
\label{intro-sec}

A common assumption in work on collaborative problem solving is that
interaction should be efficient. When language is the
mode of interaction, the measure of efficiency has been, in the main,
the number of utterances required to complete the dialogue
\cite{Chapanis72}.  One problem with this efficiency measure is
that it ignores the cognitive effort required by resource limited
agents in collaborative problem solving.  Another problem is that an
utterance-based efficiency measure shows no sensitivity to the
required quality and robustness of the problem solution.

Cognitive effort is involved in processes such as making inferences
and swapping items from long term memory into working memory.
When agents have limited working memory, then only a limited
number of items can be {\sc salient}, i.e. accessible in working
memory. Since other processes, e.g. inference, operate on salient
items, an inference process may require the cognitive effort involved
with retrieving items from long term memory, in addition to the effort
involved with reasoning itself.

The required quality and robustness of the problem solution often
determines exactly how much cognitive effort is required.  This means
that a resource-limited agent may do well on some tasks but not on
others \cite{NormanBobrow75}. For example, consider constraint-based
tasks where it is difficult for an agent to simultaneously keep all
the constraints in mind, or inference-based tasks that require a long
deductive chain or the retrieval of multiple premises, where an agent
may not be able to simultaneously access all of the required premises.

Furthermore, contrary to the efficiency hypothesis, analyses of
problem-solving dialogues shows that human agents in dialogue engage
in apparently inefficient conversational behavior. For example,
naturally-occurring dialogues often include utterances that realize
facts that are already mutually believed, or that would be mutually
believed if agents were logically omniscient
\cite{PHW82,FJW86,Walker93c}.  Consider \ex{1}-26a, which repeats
information given in \ex{1}-20 $\ldots$ \ex{1}-23:

\begin{small}
\eenumsentence
{\item[] (20) H: Right. The maximum amount of credit that you will be
able to get will be 400 {\it that they will be able to get will be 400
dollars on their tax return} \\ (21) C: {\it 400 dollars for the whole
year}?  \\ (22) H: {\it Yeah it'll be 20\%} \\ (23) C: {\it um hm} \\
(24) H: Now if indeed they pay the \$2000 to your wife, that's great.
\\ (25) C: um hm \\ (26a) H: SO WE HAVE 400 DOLLARS. \\ 
(26b) Now as far as you are concerned, that could cost you more.....
\label{so-400-examp}
}
\end{small}

Utterances such as \ex{0}-26a, that repeat, paraphrase or make
inferences explicit, are collectively called {\sc informationally
redundant utterances}, IRUs. In \ex{0}, the utterances that originally
added the belief that {\it they will get 400 dollars}  to the context
are in {\it italics} and the IRU is given in CAPS.

About 12\% of the utterances in a corpus of 55 naturally-occurring
problem-solving dialogues were IRUs \cite{Walker93c}, but the
occurrence of IRUs contradicts fundamental assumptions of many
theories of communication \cite{AP80}, {\it inter alia}. The
hypothesis that is investigated in this paper is that IRUs such as
\ex{0}-26a are related to agents' limited attentional and inferential
capacity and reflect the fact that beliefs must be salient to be used
in deliberation and inference.\footnote{The type of IRU in \ex{0}-26a
represents the Attention class of IRUs; Attitude and Consequence IRUs
are discussed elsewhere \cite{Walker92a,Walker93c}.} Hence apparently
redundant information serves an important cognitive function.

In order to test the hypothesized relationship of communicative
strategies to agents' resource limits we developed a test-bed
environment, Design-World, in which we vary task requirements, agents'
resources and communicative strategies. Our artificial agents are
based on a cognitive model of attention and memory.  Our experimental
results show that communicative strategies that incorporate IRUs can
help resource-limited cognitive agents coordinate, limit processing,
and improve the quality and robustness of the problem solution.  We
will show that the task determines whether a communicative strategy is
beneficial, depending on how the task is defined in terms of fault
intolerance and the level of belief coordination required.

\section{Design-World Task and Agent Architecture}
\label{dw-sec}

\begin{figure}[htb]
\centerline{\psfig{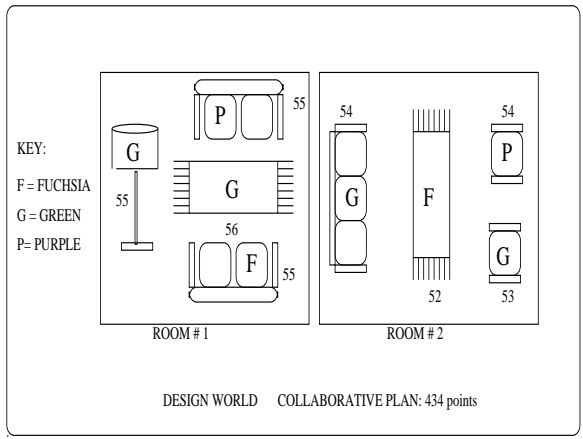}}
\caption{Potential Final State for Design-World Task: A
Collaborative Plan Achieved by the Dialogue}
\label{final-state-fig}
\end{figure}

The Design-World task consists of two agents who carry out a dialogue
in order to come to an agreement on a furniture layout design for a
two room house \cite{WGR93}.  Figure \ref{final-state-fig} shows a
potential final plan constructed as a result of a dialogue.  The
agents' shared intention is to design the house, which requires two
subparts of designing room-1 (the study) and designing room-2 (the
living room).  A room design consists of four intentions to {\sc put}
a furniture item into the room. Each furniture item has a color and
point value, which provides the basis for calculating the utility of a
{\sc put-act} involving that furniture item.  Agents start with
private beliefs about the furniture items they have and their colors.
Beliefs about which furniture items exist and how many points they are
worth are mutual.

\begin{figure}[htb]
\centerline{\psfig{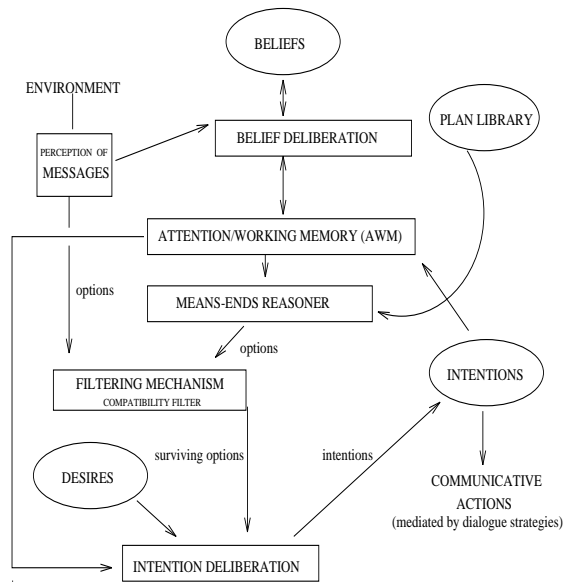}}
\caption{Design-World version of the IRMA Agent Architecture for
Resource-Bounded Agents with Limited Attention (AWM)}
\label{irma-fig}
\end{figure}

The agent architecture for deliberation and means-end reasoning is
based on the IRMA architecture, also used in the TileWorld simulation
environment \cite{BIP88,PollackRinguette90}, with the addition of a
model of limited Attention/Working memory, AWM.  See figure
\ref{irma-fig}.

The Attention/Working Memory model, AWM, is adapted from
\cite{Landauer75}. While the AWM model is extremely simple, Landauer
showed that it could be parameterized to fit many empirical results on
human memory and learning \cite{Baddeley86}.  AWM
consists of a three dimensional space in which propositions acquired
from perceiving the world are stored in chronological sequence
according to the location of a moving memory pointer.  The sequence of
memory loci used for storage constitutes a random walk through memory
with each loci a short distance from the previous one.  If items are
encountered multiple times, they are stored multiple times
\cite{HintzmannBlock71}.

When an agent retrieves items from memory, search starts from the
current pointer location and spreads out in a spherical fashion.
Search is restricted to a particular search radius: radius is defined
in Hamming distance.  For example if the current memory pointer loci
is (0 0 0), the loci distance 1 away would be (0 1 0) (0 -1 0) (0 0 1)
(0 0 -1) (-1 0 0) (1 0 0). The actual locations are calculated modulo
the memory size. The limit on the search radius defines the capacity
of attention/working memory and hence defines which stored beliefs and
intentions are {\sc salient}.

The radius of the search sphere in the AWM model is used as the
parameter for Design-World agents' resource-bound on attentional
capacity.  In the experiments below, memory is 16x16x16 and the radius
parameter varies between 1 and 16, where AWM of 1 gives severely
attention limited agents and AWM of 16 means that everything an agent
knows is salient.

The advantages of the AWM model is that it was shown to reproduce, in
simulation, many results on human memory and learning.  Because search
starts from the current pointer location, items that have been stored
most recently are more likely to be retrieved, predicting recency
effects \cite{Baddeley86}.  Because items that are stored in multiple
locations are more likely to be retrieved, the model predicts
frequency effects \cite{Landauer75}.  Because items are
stored in chronological sequence, the model produces natural
associativity effects \cite{AB73}.  Because deliberation and means-end
reasoning can only operate on salient beliefs, limited attention
produces a concomitant inferential limitation, i.e. if a belief is not
salient it cannot be used in deliberation or means-end-reasoning.
This means that mistakes that agents make in their planning process
have a plausible cognitive basis. Agents can both
fail to access a belief that would allow them to produce an optimal
plan, as well as make a mistake in planning if a belief about how the
world has changed as a result of planning is not salient.

\section{Design-World Communicative Strategies}
\label{dial-strat-sec}

\begin{figure}[htb]
\centerline{\psfig{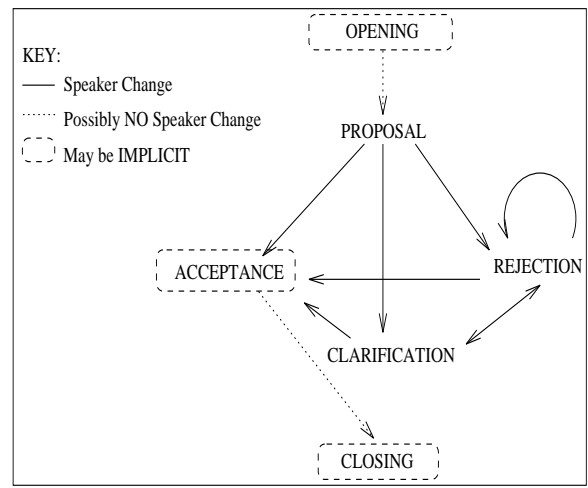}}
\caption{Discourse Actions for the Design-World Task}
\label{dial-stat-fig}
\end{figure}

A {\sc communicative strategy} is a strategy for communicating with
another agent, which varies according to the agents' initiative,
amount of information about the task, degree of resource-bounds, and
communication style \cite{WW90,Carletta92,Cawsey92a,Guinn93}.
Design-World agents communicate with an artificial language whose
primitive communicative acts are {\sc propose, accept, reject, say}.
These primitive acts can be composed to produce higher level discourse
acts such as {\sc proposals, acceptances, rejections, openings} and
{\sc closings} \cite{Walker93c}. See figure \ref{dial-stat-fig}.

A discourse act may be left implicit, or may be varied to consist of
one or more communicative acts.  Discourse acts are different from
actions on the environment because they are actions whose intended
effect is a change in the other agent's mental state. Because the
other agent is an active intelligence, it is possible for it to
supplement an underspecified discourse action with its own processing.
The variation in the degree of explicitness of a discourse act is the
basis of agents' communicative strategies.  Here we will compare three
communicative strategies: (1) All-Implicit; (2) Close-Consequence; and
(3) Explicit-Warrant.

The All-Implicit strategy is a `bare bones' strategy, exemplified by
the partial dialogue in \ex{1}.  In \ex{1} each utterance is shown
both as a gloss in {\it italics}, and in the artificial language that
the agents communicate with.

\begin{small}
\eenumsentence
{
\item[1:]
BILL: {\it Then, let's put the green rug in the study.}\\
(propose agent-bill agent-kim option-43:  put-act (agent-bill green
rug room-1))

\item[2:]
KIM: {\it Then, let's put the green lamp in the study.} \\
(propose agent-kim agent-bill option-61:  put-act (agent-kim
green lamp room-1))

\item[3:]
BILL: {\it No, instead let's put the green couch in the study.} \\
(reject agent-bill agent-kim option-75:  put-act (agent-bill green couch
room-1))

.....
}
\end{small}

In Design-World, unlike TileWorld, an option that is generated via
means-end reasoning or from proposals of other agents only becomes an
intention if it is {\sc accepted} by both agents. See figure
\ref{dial-stat-fig}. In dialogue \ex{0}, Bill makes a proposal in 1,
and then Kim implicitly accepts this proposal with a new proposal in
2.  In \ex{0}-3 Bill rejects Kim's proposal and makes a
counter-proposal.

The content of communicative acts are beliefs and (potential)
intentions. Dialogue \ex{0} illustrates part of the cycle for
achieving a {\sc design-house} plan: (1) individual agents {\sc
means-end reason} about options in the domain; (2) individual agents
{\sc deliberate} about which options are preferable; (3) then agents
make {\sc proposals} to other agents, based on the options identified
in a reasoning cycle, about actions that {\sc contribute} to the
satisfaction of their intentions; (4) then these proposals are {\sc
accepted} or {\sc rejected} by the other agent, or
acceptance/rejection is postponed by {\sc asking} for more
information.  See figure \ref{irma-fig}. Deliberating whether to
accept or reject a proposal is based on beliefs about the proposed
action's utility \cite{Doyle92}.

Agents parameterized with the All-Implicit strategy do not include
IRUs in any discourse act or produce any discourse acts labelled as
potentially implicit in figure \ref{dial-stat-fig}.  Agents
parameterized with the Close-Consequence and Explicit-Warrant
strategies include IRUs at dialogue segment closings and in proposals.

In dialogue \ex{1} agent CLC uses the Close-Consequence strategy.  CLC
makes explicit {\sc closing} statements, such as \ex{1}-2, on the
completion of the intention associated with a discourse segment.
CLC's {\sc closing} discourse act also includes IRUs as in \ex{1}-3;
CLC makes the inference explicit that since they have agreed on
putting the green rug in the study, Bill no longer has the green rug
(act-effect inference).

\begin{small}
\eenumsentence
{
\item[1:]
BILL: {\it Then, let's put the green rug in the study.} \\
(propose agent-bill agent-clc option-30: put-act (agent-bill green
rug room-1))

\item[2:]
CLC: {\it So, we've agreed to put the green rug in the study.} \\
(close agent-clc agent-bill intended-30: put-act (agent-bill green
rug room-1))

\item[3:]
CLC: {\sc agent-bill doesn't have green rug.} \\
(say agent-clc agent-bill bel-48: has n't (agent-bill green rug))

\label{ccl-dial}
}
\end{small}

The Close-Consequence strategy of making inferences explicit at the
close of a segment is intended to parallel the naturally occurring
example in \ref{so-400-examp}.  In both cases an inference is made
explicit that follows from what has just been said, and the inference
is sequentially located at the close of a discourse segment.

The Explicit-Warrant strategy varies the proposal discourse act by
including {\sc warrant} IRUs in each proposal.  In general a {\sc
warrant} for an intention is a reason for adopting the intention, and
here {\sc warrants} are the score propositions that give the utility
of the proposal, which are mutually believed at the outset of the
dialogues.  In \ex{1}, the {\sc warrant} IRU is in {\sc caps}.


\begin{small}
\eenumsentence
{
\item[1:]
IEI: {\sc  Putting in the green rug is worth  56} \\
(say agent-iei agent-iei2 bel-265: score  (option-202:  put-act
(agent-bill green rug room-1) 56))

\item[2:]
IEI: {\it Then, let's put the green rug in the study.} \\
(propose agent-iei agent-iei2 option-202:  put-act (agent-bill green rug
room-1))
}
\end{small}

Since warrants are used by the other agent in deliberation, the
Explicit-Warrant strategy can save the other agent the processing
involved with determining which facts are relevant for deliberation
and retrieving them from memory.   The Explicit-Warrant strategy
also occurs in natural dialogues \cite{Walker93c}.

\section{Design World Task Variations}
\label{task-def-effect-sec}

Design-World supports the parameterization of the task so that it can
be made more difficult to perform by making greater processing demands
on the agents.  These task variations will be shown to interact with
variations in communicative strategies and attentional capacity in
section \ref{results-sec}.

\begin{figure}[htb]
\centerline{\psfig{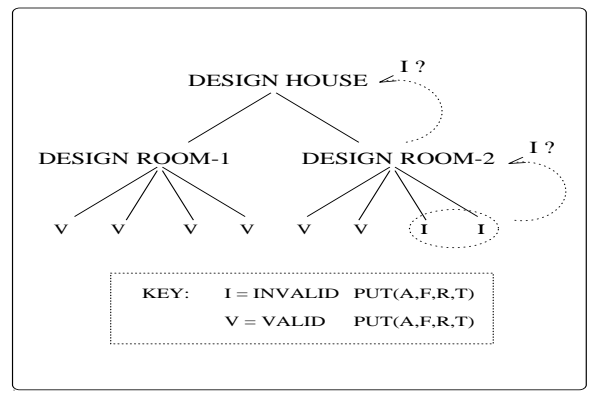}}
\caption{Evaluating Task Invalids: for some tasks invalid steps
invalidate the whole plan. }
\label{task-eval-inval-fig}
\end{figure}

\begin{figure}[htb]
\centerline{\psfig{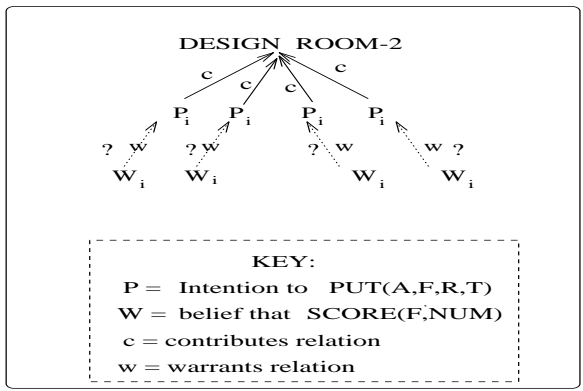}}
\label{task-eval-fig}
\caption{Tasks can differ as to the level of mutual belief required.
Some tasks require that the {\sc warrant} W, a reason for
doing P, is mutually believed and others don't.}
\end{figure}

\subsubsection{Standard Task}

The Standard task is defined so that the {\sc raw score} that agents
achieve for a {\sc design-house} plan, constructed via the dialogue,
is the sum of all the furniture items for each valid step in their
plan. The point values for invalid steps in the plan are simply
subtracted from the score so that agents are not heavily penalized for
making mistakes.

\subsubsection{Zero Invalids Task }

The Zero-Invalids Task is a fault-intolerant version of the task in
which any invalid intention invalidates the whole plan. In general,
the effect of making a mistake in a plan depends on how interdependent
different subparts of the problem solution are.\footnote{Contrast
aircraft scheduling with furnishing a room.} Figure
\ref{task-eval-inval-fig} shows the choices for the effect of invalid
steps for the Design-World task.  The score for invalid steps
(mistakes) can just be subtracted out; this is how the Standard task
is defined.  Alternately, invalid steps can propagate up so that an
invalid {\sc put-act} means that the Design-Room plan is invalid.
Finally, mistakes can completely propagate so that the Design-House
plan is invalid if one step is invalid, as in the Zero-Invalids task.

\subsubsection{Zero NonMatching Beliefs Task }

The Zero-Nonmatching-Beliefs task is designed to investigate the
effect of the level of agreement that agents must achieve.   Figure
\ref{task-eval-fig} illustrates different degrees of agreeing in a
collaborative task, e.g.  agents may agree on the actions to be done,
but not agree on the {\bf reasons} for intending that
action.\footnote{Consider a union/ management negotiation where each
party has different reasons for any agreement.} The
Zero-NonMatching-Beliefs task is defined so that a {\sc warrant} W, a
reason for doing P, must be mutually supposed.

\section{Experimental Results}
\label{results-sec}

\begin{figure}[htb]
\centerline{\psfig{figure=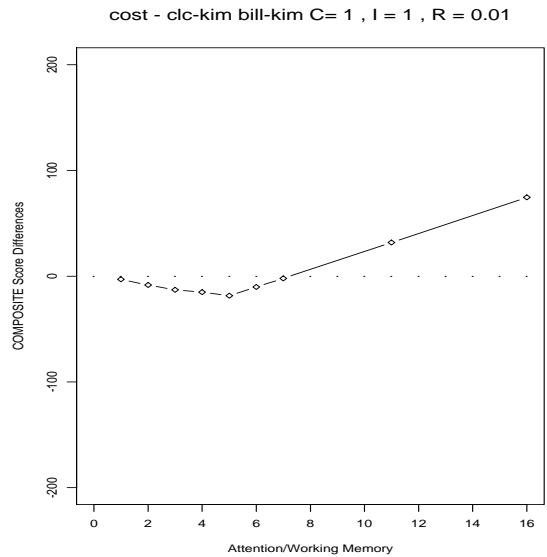,height=3.0in,width=3.0in}}
\caption{Close-Consequence can be detrimental in the Standard Task. Strategy 1
is the combination of an All-Implicit
agent with a Close-Consequence agent and Strategy 2
is two All-Implicit agents,
Task  = Standard, commcost = 1, infcost = 1, retcost = .01}
\label{cost-clc-kim-fig}
\end{figure}

We wish to evaluate the relative benefits of the communicative
strategies in various tasks for a range of resource limits.  In
section \ref{task-def-effect-sec} we defined an objective performance
measure for the {\sc design-house} plan for each task variation. We
must also take cognitive costs into account.  Because cognitive effort
can vary according to the communication situation and the agent
architecture, performance evaluation introduces three additional
parameters: (1) {\sc commcost}: cost of sending a message; (2) {\sc
infcost}: cost of inference; and (3) {\sc retcost}: cost of retrieval
from memory:

\begin{quote}
\begin{tabular}{l}
{\sc performance} $=$  \\
 Task Defined {\sc raw score} \\
 -- ({\sc commcost} $\times$ total messages)\\
 -- ({\sc infcost} $\times$ total inferences)\\
 -- ({\sc retcost} $\times$ total retrievals)
\end{tabular}
\end{quote}



We simulate 100 dialogues at each parameter setting and calculate the
normalized performance distributions for each sample run.  In the
results to follow, {\sc commcost}, {\sc infcost} and {\sc retcost} are
fixed at 1,1, .01 respectively, and the parameters that are varied are
(1) communication strategy; (2) task definition; and (3) AWM
settings.\footnote{See \cite{Walker93c,Walker95} for results related
to varying the relative cost of retrieval, inference and
communication.} Differences in the performance distributions for each
set of parameters are evaluated for significance over the 100
dialogues using the Kolmogorov-Smirnov (KS) two sample test
\cite{Siegel56}.

A strategy A is defined to be {\sc beneficial} as compared to a
strategy B, for a set of fixed parameter settings, if the difference
in distributions using the Kolmogorov-Smirnov two sample test is
significant at p $<$ .05, in the positive direction, for two or more
AWM settings.  A strategy is {\sc detrimental} if the differences go
in the negative direction.  Strategies may be neither {\sc beneficial}
or {\sc detrimental}, since there may be no difference between two
strategies.

A {\sc difference plot} such as that in figure
\ref{cost-clc-kim-fig} will be used to summarize a comparison of
strategy 1 and strategy 2. In the comparisons below, strategy 1 is
either Close-Consequence or Explicit-Warrant and strategy 2 is the
All-Implicit strategy.  {\bf Differences} in performance between two
strategies are plotted on the Y-axis against AWM parameter settings on
the X-axis.  Each point in the plot represents the difference in the
means of 100 runs of each strategy at a particular AWM setting.  These
plots summarize the information from 18 performance distributions
(1800 simulated dialogues).  Every simulation run varies the AWM
radius from 1 to 16 to test whether a strategy only has an effect at
particular AWM settings.  If the plot is above the dotted line for 2
or more AWM settings, then strategy 1 may be {\sc beneficial},
depending on whether the differences are significant.\footnote{Visual
difference in means and distributional differences need not be
correlated, however KS significance values will be given with each
figure, and difference plots are much more concise than actual
distributions.}

In the reminder of this section, we first compare within strategy, for
each task definition and show that whether or not a strategy is
beneficial depends on the task. Then we compare across strategies for
a particular task, showing that the interaction of the strategy and
task varies according to the strategy.  The comparisons will show that
what counts as a good collaborative strategy depends on cognitive
limits on attention and the definition of success for the task.

\subsection{Close Consequence}

\begin{figure}[htb]
\centerline{\psfig{figure=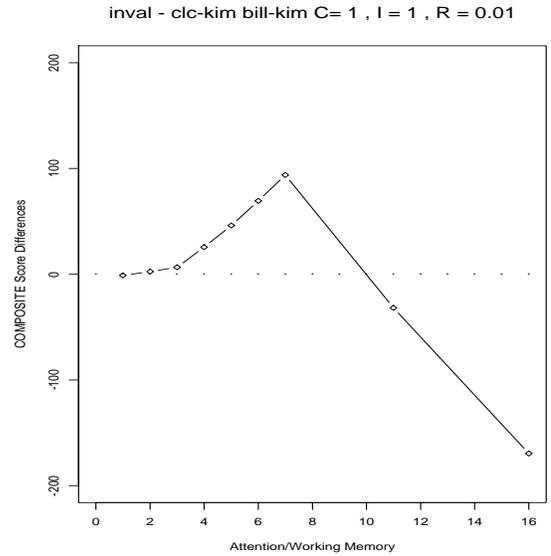,height=3.0in,width=3.0in}}
\caption{Close Consequence is beneficial for Zero-Invalids Task. Strategy 1 is
the combination of an All-Implicit
agent with a Close-Consequence agent and Strategy 2 is two
All-Implicit agents, Task  = Zero-Invalid, commcost = 1,
infcost = 1, retcost = .01}
\label{clc-inval-fig}
\end{figure}

The difference plot in figure \ref{cost-clc-kim-fig} shows that
Close-Consequence is {\sc detrimental} in the Standard task at AWM of
1 $\ldots$ 5 (KS $>$ 0.19, p $<$ .05).

In contrast, if the task is the fault-intolerant Zero-Invalids task,
then the Close-Consequence strategy is {\sc beneficial}.  Figure
\ref{clc-inval-fig} demonstrates that strategies which include
Consequence IRUs can increase the robustness of the planning process
by decreasing the frequency with which agents make mistakes (KS for
AWM of 3 to 6 $>$ .19, p $<$ .05). This is a direct result of {\bf
rehearsing} the act-effect inferences, making it unlikely that
attention-limited agents will forget that they have already used a
furniture item.

\begin{figure}[htb]
\centerline{\psfig{figure=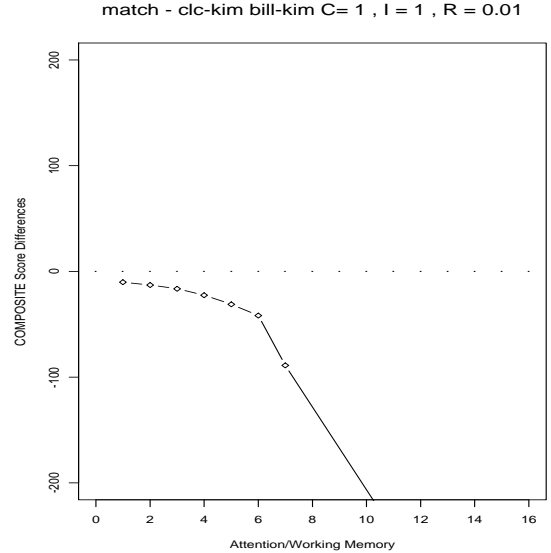,height=3.0in,width=3.0in}}
\caption{Close-Consequence is detrimental for
Zero-Nonmatching-Beliefs Task. Strategy 1 is the combination of an All-Implicit
agent with a Close-Consequence agent and Strategy 2
is two All-Implicit agents,
Task  =  Zero-Nonmatching-Beliefs, commcost = 1, infcost = 1, retcost = .01}
\label{clc-nmb-fig}
\end{figure}

Figure \ref{clc-nmb-fig} shows that the Close-Consequence
strategy is detrimental when the task requires agents to achieve
matching beliefs on the {\sc warrants} for their intentions (KS 1,3)
$>$ 0.3, p $<$ .01).  This is because IRUs displace other facts from
AWM. In this case agents forget the scores of furniture pieces under
consideration, which are the warrants for their intentions. Thus here,
as elsewhere, we see that IRUs can be detrimental by making agents
forget critical information.

\subsection{Explicit Warrant}

\begin{figure}[htb]
\centerline{\psfig{figure=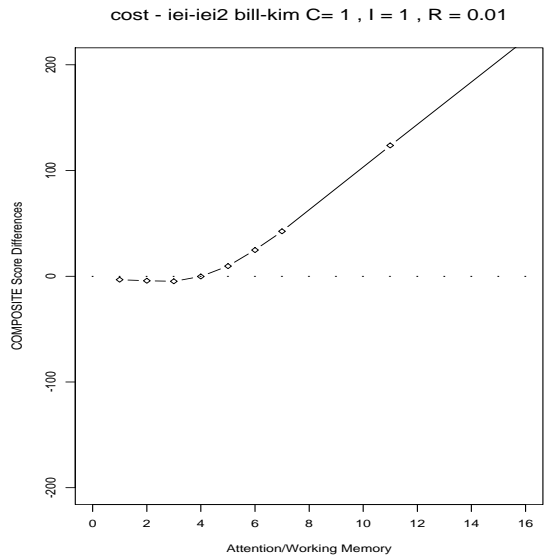,height=3.0in,width=3.0in}}
\caption{Explicit-Warrant saves Retrieval costs: Strategy 1 is
two Explicit-Warrant agents and strategy 2 is two All-Implicit agents:
Task = Standard, commcost = 1, infcost = 1,
retcost = .01}
\label{ret-iei-fig}
\end{figure}

Figure \ref{ret-iei-fig} shows that Explicit-Warrant is beneficial in
the Standard task at AWM values of 3 and above.  Here, the
scores improve  because the beliefs necessary for deliberating
the proposal are made available in the current context with each
proposal (KS for AWM of 3 and above $>$ .23, p $<$ .01), so that
agents don't have to search memory for them.  At AWM parameter
settings of 16, where agents can search a huge belief space for
beliefs to be used as warrants, the saving in processing time is
substantial.

When the task is Zero-Invalid (no figure due to space), the
benefits of the Explicit-Warrant strategy are dampened from the
benefits of the Standard task, because Explicit-Warrant does nothing
to address the reasons for agents making mistakes.  In comparison with
the All-Implicit strategy, it is detrimental at AWM of 1 and 2, but is
still beneficial at AWM of 5,6,7, and 11.



\begin{figure}[htb]
\centerline{\psfig{figure=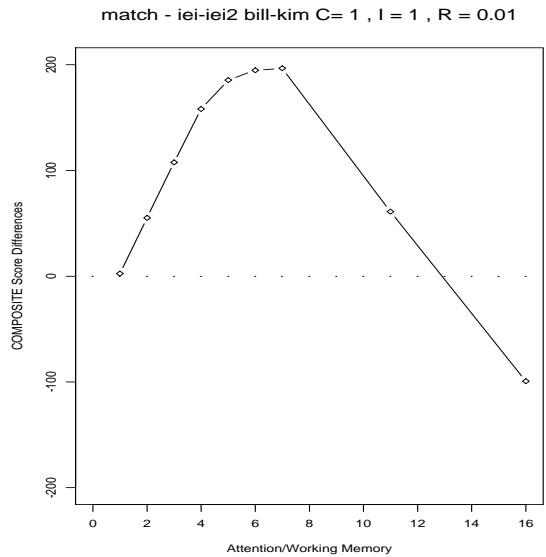,height=3.0in,width=3.0in}}
\caption{Explicit-Warrant is beneficial for Zero-NonMatching-Beliefs
Task: Strategy 1 is two Explicit-Warrant agents and strategy 2 is two
All-Implicit agents: Task = Zero-Nonmatching-Beliefs, commcost = 1,
infcost = 1, retcost = .01}
\label{iei-nmb-fig}
\end{figure}

In contrast to Close-Consequence, the Explicit-Warrant strategy is
highly beneficial when the task is Zero-NonMatching-Beliefs, see
figure \ref{iei-nmb-fig} (KS $>$ .23 for AWM from 2 to 11, p $<$ .01).
When agents must agree on the warrants underlying their intentions,
including these warrants with proposals is a good strategy even if the
agent already knows the warrants. This is due to agents' resource
limits, which means that retrieval is indeterminate and that there are
costs associated with retrieving warrants from memory. At high AWM the
differences between the two strategies are small.

\section{Related Work}

Design-World was inspired by the TileWorld simulation environment: a
rapidly changing robot world in which an artificial agent attempts to
optimize reasoning and planning \cite{PollackRinguette90,HPC93}.
TileWorld is a single agent world in which the agent interacts with
its environment, rather than with another agent.  Design-World uses
similar methods to test a theory of the effect of resource limits on
communicative behavior between two agents.

The belief reasoning mechanism of Design-World agents was
informed by the theory of belief revision and the multi-agent
simulation environment developed in the Automated Librarian project
\cite{Galliers91b,Cawsey92a}. The communicative acts and discourse
acts used by Design-World agents are similar to those used in
\cite{Carletta92,Cawsey92a,Sidner92,Stein93}.

Design-World is also based on the method used in Carletta's JAM
simulation for the Edinburgh Map-Task \cite{Carletta92}.  JAM is based
on the Map-Task Dialogue corpus, where the goal of the task is for the
planning agent, the instructor, to instruct the reactive agent, the
instructee, how to get from one place to another on the map.  JAM
focuses on efficient strategies for recovery from error and
parametrizes agents according to their communicative and error
recovery strategies.  Given good error recovery strategies, Carletta
argues that `high risk' strategies are more efficient, where
efficiency is a measure of the number of utterances in the dialogue.
While the focus here is different, we have shown that that the number
of utterances is just one parameter for evaluating performance, and
that the task definition determines when strategies are effective.

\section{Conclusion}

In this paper we showed that collaborative communicative behavior
cannot be defined in the abstract: what counts as collaborative
depends on the task, and the definition of success in the task.  We
used two empirical methods to support our argument: corpus based
analysis and experimentation in Design-World. The methods and the
focus of this work are novel; previous work on resource limited agents
has not examined the role of communicative strategies in multi-agent
interaction whereas work on communication has not considered the
effects of resource limits.

We showed that strategies that are inefficient under assumptions of
perfect reasoners with unlimited attention and retrieval are effective
with resource limited agents.  Furthermore, different tasks make
different cognitive demands, and place different requirements on
agents' collaborative behavior. Tasks which require a high level of
belief coordination can benefit from communicative strategies that
include redundancy. Fault intolerant tasks benefit from redundancy
for rehearsing the effects of actions.

Because the communicative strategies that we tested were based on a
corpus analysis of human human financial advice dialogues and because
variations in the Design-World task were parametrized, we believe the
results presented here may be domain independent, though clearly more
research is needed.

Here we fixed the parameters for the cost of communication, inference
and retrieval, only discussed a few of the implemented discourse
strategies, and didn't discuss Design-World parameters that increase
the inferential complexity of the task and that limit inferential
processing.  Elsewhere we show that: (1) when retrieval is free or
when communication cost is high, that the Explicit-Warrant strategy is
detrimental at low AWM \cite{Walker93c}; (2) some IRU
strategies are only beneficial when inferential complexity is higher
than in the Standard Task \cite{Walker93}; (3) IRUs that
make inferences explicit can help inference limited agents perform as
well as logically omniscient ones \cite{Walker95}.

One ramification of the results presented here is that experimental
environments for testing agent architectures should support task
variation \cite{PollackRinguette90,HPC93}. Furthermore the task
variation should test aspects of the interaction of the agents
involved. These results also inform the design of multi-agent problem
solving systems and for systems for teaching, advice and explanation.

\end{document}